\documentclass[a4paper,conference]{IEEEtran}
 
\usepackage[T1]{fontenc}
\usepackage[utf8]{inputenc}
\usepackage{amsmath}
\usepackage{amsfonts}
\usepackage{amssymb}
\usepackage{float}
\usepackage[figuresright]{rotating}
\usepackage{pifont}
\usepackage{microtype}
\usepackage{subfig}
\usepackage{cite}
\usepackage{multirow}
\usepackage{booktabs}
\usepackage{tabularx}
\usepackage[group-separator={,}]{siunitx}
\usepackage{xcolor}
\usepackage{enumitem}
\usepackage{mathtools}
\usepackage{graphicx}
\usepackage[outdir=figs/]{epstopdf}

\usepackage{url}
\usepackage{balance}
\usepackage[bookmarks=false]{hyperref}
 
\hypersetup{
  linkcolor  = red!50!black,
  citecolor  = blue!80!black,
  urlcolor   = violet!85!black,
  colorlinks = true,
}
 
\makeatletter
\renewcommand{\@biblabel}[1]{[#1]\hfill}
\makeatother
 
\let\oldtimes\times
\def\times{{\mkern1mu\oldtimes\mkern1mu}}

\DeclarePairedDelimiter{\abs}{\lvert}{\rvert}
 
\usepackage{xspace}
\makeatletter
\DeclareRobustCommand\onedot{\futurelet\@let@token\@onedot}
\def\@onedot{\ifx\@let@token.\else.\null\fi\xspace}

\def\etal{et al\onedot}
\makeatother

\begin{document}
\tolerance=999
\sloppy

\title{Top-DB-Net: Top DropBlock for Activation \\ Enhancement in Person Re-Identification}
 
\author{
  \IEEEauthorblockN{
  Rodolfo Quispe\IEEEauthorrefmark{1}\IEEEauthorrefmark{2}
  \hspace*{0.2cm} Helio Pedrini\IEEEauthorrefmark{2}\IEEEauthorrefmark{3B}
  }
  \IEEEauthorblockA{\IEEEauthorrefmark{1}Microsoft Corp., One Microsoft Way, Redmond, WA, USA, 98052-6399}
  \IEEEauthorblockA{\IEEEauthorrefmark{2}Institute of Computing, University of Campinas, Campinas, SP, Brazil, 13083-852}
  \IEEEauthorblockA{\IEEEauthorrefmark{3}
  Email: helio@ic.unicamp.br}
}
 
\maketitle
 
\begin{abstract}
Person Re-Identification is a challenging task that aims to retrieve all instances of a query image across a system of non-overlapping cameras. Due to the various extreme changes of view, it is common that local regions that could be used to match people are suppressed, which leads to a scenario where approaches have to evaluate the similarity of images based on less informative regions. In this work, we introduce the Top-DB-Net, a method based on Top DropBlock that pushes the network to learn to focus on the scene foreground, with special emphasis on the most task-relevant regions and, at the same time, encodes low informative regions to provide high discriminability. The Top-DB-Net is composed of three streams: (i) a global stream encodes rich image information from a backbone, (ii) the Top DropBlock stream encourages the backbone to encode low informative regions with high discriminative features, and (iii) a regularization stream helps to deal with the noise created by the dropping process of the second stream, when testing the first two streams are used. Vast experiments on three challenging datasets show the capabilities of our approach against state-of-the-art methods. Qualitative results demonstrate that our method exhibits better activation maps focusing on reliable parts of the input images. The source code is available at: \href{https://github.com/RQuispeC/top-dropblock}{\url{https://github.com/RQuispeC/top-dropblock}}.
\end{abstract}
 
\IEEEpeerreviewmaketitle
 
\section{Introduction}
 
Person Re-Identification (ReID) aims to match all the instances of the same person across a system of non-overlapping cameras. This is a challenging task due to extreme view-point changes and occlusions. It has various applications in surveillance systems and it has gained a lot of popularity in the context of computer vision, where new scenarios of this task have been developed recently~\cite{zhou2018vehicle,liu2016large,zhang2019mvb}.
 
Numerous approaches have been proposed using person-related information, such as pose and body parts~\cite{quispe2019improved,kumar2017pose,zheng2019pose,li2017learning,cheng2016person}. However, ReID datasets only provide ID labels. Thus, these methods rely on other datasets proposed for related tasks during the training. This dependency introduces further errors in predictions and motivates the creation of general methods that do not learn from outer information.
 
In this paper, we introduce the Top DropBlock Network (Top-DB-Net) for the ReID problem. Top-DB-Net is designed to further push networks to focus on task-relevant regions and encode low informative regions with discriminative features.
 
Our method is based on three streams consisting of (i) a classic global stream as most of the state-of-the-art methods~\cite{quispe2019improved,kumar2017pose,zheng2019pose,li2017learning,cheng2016person, dai2019batch,luo2019bag}, (ii) a second stream drops\footnote{We use the terms {\it remove} and {\it drop} interchangeably to indicate that a tensor region has been zeroed out.} most activated horizontal stripes of feature tensors to enhance activation in task-discriminative regions and improve encoding of low informative regions, and (iii) a third stream regularizes the second stream avoiding that noise generated by dropping features degrades the final results.
 
\begin{figure}[!htb]
  \captionsetup[subfigure]{labelformat=empty}
  \centering
  \subfloat{\includegraphics[width=1.4cm, height=3.0cm]{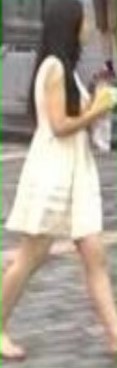}} \hspace*{0.1cm}
  \subfloat{\includegraphics[width=1.4cm, height=3.0cm]{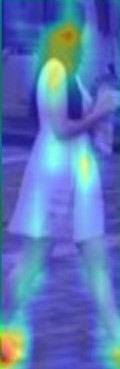}} \hspace*{0.1cm}
  \subfloat{\includegraphics[width=1.4cm, height=3.0cm]{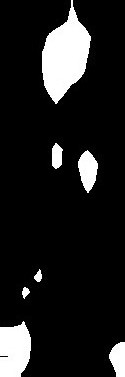}} \hspace*{0.1cm}
  \subfloat{\includegraphics[width=1.4cm, height=3.0cm]{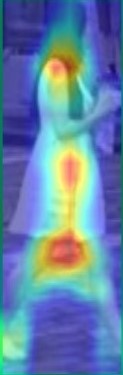}}\hspace*{0.1cm}
  \subfloat{\includegraphics[width=1.4cm, height=3.0cm]{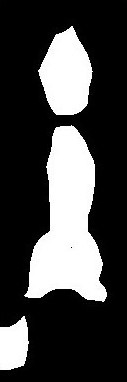}} \\
  \subfloat[input image]{\includegraphics[width=1.4cm, height=3.0cm]{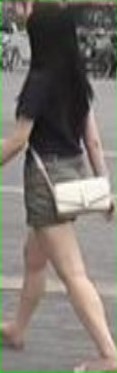}} \hspace*{0.1cm}
  \subfloat[baseline]{\includegraphics[width=1.4cm, height=3.0cm]{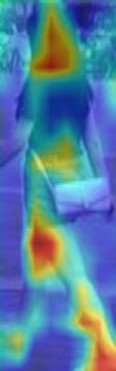}} \hspace*{0.1cm}
  \subfloat[baseline]{\includegraphics[width=1.4cm, height=3.0cm]{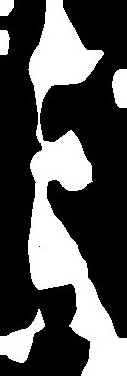}} \hspace*{0.1cm}
  \subfloat[ours]{\includegraphics[width=1.4cm, height=3.0cm]{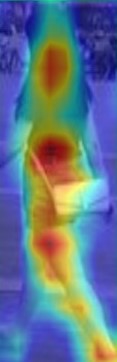}} \hspace*{0.1cm}
  \subfloat[ours]{\includegraphics[width=1.4cm, height=3.0cm]{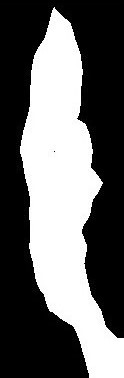}}
  \caption{Comparison of activation maps generated by the proposed method and a baseline~\cite{dai2019batch}. The first column shows the input images, the second and fourth columns present the activation maps that overlap the input images, and the third and fifth columns show a mask generated by thresholding the activation maps.}
  \label{fig:comp-activation-baseline}
  \end{figure}

As a result of our proposed method, we can observe in Figure~\ref{fig:comp-activation-baseline} that the activation maps~\cite{zagoruyko2016paying} generated by our baseline, focus both on body parts and background, whereas Top-DB-Net focus consistently on the body with stronger activation to discriminative regions.
 
\begin{figure}[!htb]
  \captionsetup[subfigure]{labelformat=empty}
  \centering
  \subfloat[Input image]{\includegraphics[width=1.8cm, height=3.2cm]{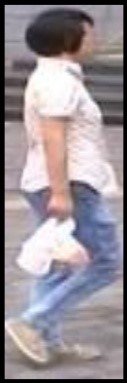}} \hspace*{0.2cm}
  \subfloat[Activation]{\includegraphics[width=1.8cm, height=3.2cm]{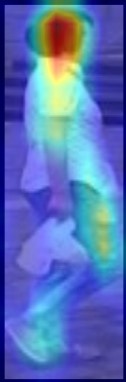}} \hspace*{0.2cm}
  \subfloat[BDB drop mask]{\includegraphics[width=1.8cm, height=3.2cm]{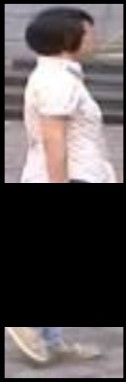}} \hspace*{0.2cm}
  \subfloat[Our drop mask]{\includegraphics[width=1.8cm, height=3.2cm]{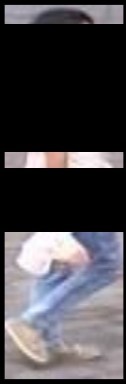}}
  \caption{Input image, its activation map after epoch 120 and drop masks. BDB creates a random drop mask, while our method creates a mask that drops most activated regions.}
  \label{fig:dropmask-comparison}
\end{figure}

Contrasting our Top-DB-Net with BDB Network~\cite{dai2019batch}, there are three differences: (i) instead of dropping random features, our method drops only features with top (the largest) activations, which stimulates the network to maintain performance using only features with inferior discriminative power (the lowest activations), (ii) rather than using the same drop mask for every feature map in the batch, our method creates an independent drop mask for every input based on its top activations, and (iii) dropping top activated features creates noise inside the second stream (Figure~\ref{fig:dropmask-comparison}), thus we introduce a third stream that forces the features before the dropping step to be still discriminative for ReID, which works as a regularizer due to the multi-task principle~\cite{goodfellow2016deep}. We use the same definition for 'batch' as Dai \etal~\cite{dai2019batch}, that is, ``group of images participating in a single loss calculation during training''. The intuition of why our implementation is better can be explained by analyzing Figure~\ref{fig:dropmask-comparison}. For an input image, we can see that the major activations are over the upper body. BDB Network~\cite{dai2019batch} creates a random drop mask that, in this case, removes the lower body during training. This would encourage the network to continue focusing on the upper body. On the other hand, our method controls which regions are being dropped and encourages the network to learn from the lower body. Our results show that this helps during the learning process (Figure~\ref{fig:activation-epoch}) and generates activation maps better spread over the foreground (Figure~\ref{fig:comp-activation-baseline}).
 
The evaluation of our proposed method is conducted through extensive experiments on three widely used datasets for ReID. We consider the BDB Network~\cite{dai2019batch} as a baseline for our work and demonstrate that our Top-DB-Net outperforms it by up to 4.7\% in the CUHK03 dataset~\cite{li2014deepReID}. Moreover, our results show competitive results against state-of-the-art approaches.
 
In Section~\ref{sec:background}, we discuss the evolution of the ReID task and review relevant related work. In Section~\ref{sec:method}, we introduce the Top-DB-Net. In Section~\ref{sec:results}, we describe our experiments and evaluate the results achieved in three challenging datasets. Finally, concluding remarks and directions for future work are presented in Section~\ref{sec:conclusions}.
 
\section{Background}
\label{sec:background}
 
The term ReID was first stated by Zajdel \etal~\cite{zajdel2005keeping} as a variation of people tracking problem. However, unlike tracking algorithms, ReID does not depend on the hypotheses of constancy. Thus, it is a more complicated problem due to considerable variations in biometric profile, position, appearance and point of view~\cite{vezzani2013people}.
 
Many initial works in ReID considered it as a classification task. This is mainly because most datasets~\cite{gray2008viewpoint} available at that time had just a few instances of each person. Because of this, various methods based on handcrafted features~\cite{corvee2010person,cheng2011custom,hirzer2012dense,hirzer2012person,leng2015person} were initially proposed.
 
With the popularization of deep learning and ReID, many datasets with larger amounts of instances per person in real-world scenarios have been made available~\cite{zheng2015scalable,li2014deepReID,zhong2017re,zheng2017unlabeled,ristani2016MTMC} and deep networks-based methods had become the standard~\cite{torchreid,zhou2019osnet}.
 
This had two side effects: (i) the most popular datasets already include a predefined training and testing split -- which helps with validation protocols and comparison between methods -- and (ii) ReID turned into a retrieval problem -- thus, measures such as Cumulated Matching Characteristics (CMC) and Mean Average Precision (mAP) are widely used.
 
Various methods proposed for ReID use specific prior related to person's nature, such as pose and body parts~\cite{quispe2019improved,kumar2017pose,zheng2019pose,li2017learning,cheng2016person}. However, labels such as segmented semantic regions and body skeleton that are necessary for these types of methods are not available in current ReID datasets. Thus, they usually leverage datasets proposed for other tasks captured in different domains, which introduces noise during training and makes the learning process more complicated.
 
On the other hand, there are methods~\cite{dai2019batch,li2018harmonious,hou2019interaction,xia2019second} that learn to encode rich information directly from the input images without relying on other types of signals. Our work follows this strategy. Most of the methods in this category use the concept of attention in their pipeline. Thus, their approaches expect networks to learn to focus on discriminative regions and encode those parts. However, assuming that the availability of consistent discriminative regions may introduce errors, since occlusions are a major problem in the context of ReID due to drastic view changes.
 
The discriminative regions that can be used to match two people may not be available in all instances, such that the approaches require to maintain performance without relying on the availability of high discriminability regions or, in other words, being able to encode richer information from less discriminative regions. In this sense, we propose a method that aims to simulate this scenario by dropping top activated (most discriminative) regions and reinforcing the network to perform ReID with only less discriminative regions available.
 
To further improve ReID performance, literature have proposed re-ranking~\cite{zhong2017re,saquib2018pose} and augmentation~\cite{zheng2019joint,zhong2018camstyle} approaches. The former methods can improve ReID results by a huge margin, which makes it unfair to compare pipelines using them against pipelines not using them. Therefore, since various state-of-the-art methods report results with and without re-ranking, our comparison to them is made separately for these two scenarios.
 
\begin{figure*}[!htb]
  \centering
  \includegraphics[width=0.9\linewidth]{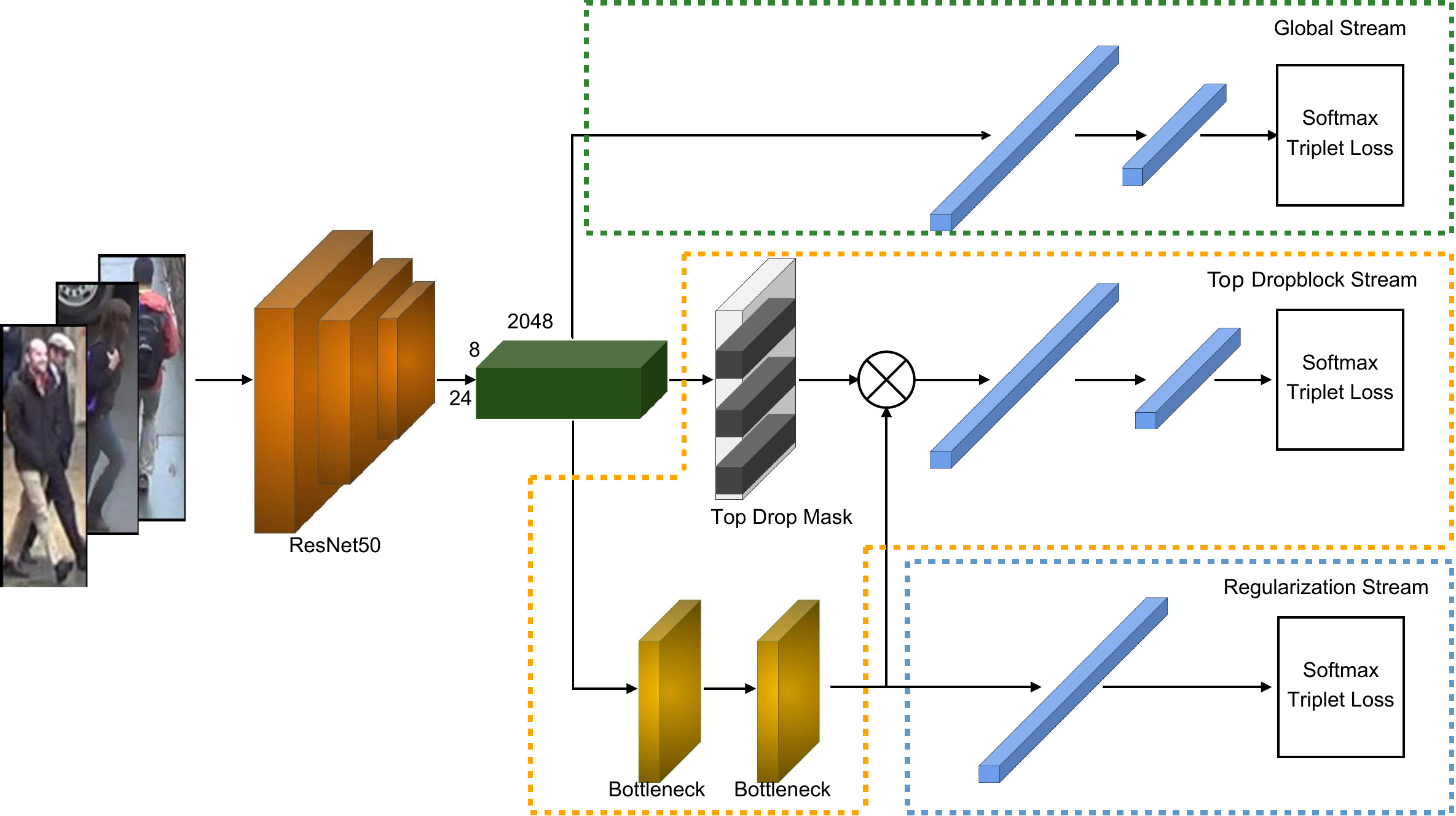}
  \caption{Proposed Top DropBlock Network (Top-DB-Net). It is composed of three streams that are able to focus on reliable parts of the input and encode low informative regions with high discriminative features for enhanced performance. It is trained using triplet loss and cross entropy. During the testing stage, the outputs of Global and Top DropBlock streams are concatenated.}
  \label{fig:architecture}
\end{figure*}
 
\section{Proposed method}
\label{sec:method}
 
This section describes our Top DropBlock Network (Top-DB-Net) for addressing the ReID problem. We first introduce our baseline based on BDB Network~\cite{dai2019batch}. Then, we present each of the three streams of our Top-DB-Net and the loss functions used to train our model. The combination of these streams leads to improvements in the final performance and activation maps.
 
\subsection{Baseline}
 
We decided to use BDB Network~\cite{dai2019batch} as the baseline for our proposal because of its similarity with our approach. BDB Network uses ResNet-50~\cite{he2016deep} as backbone as in many ReID works, however, a slight variation is made by removing the last pooling layer. Thus, a larger feature map is obtained, more specifically, with a size of 2048$\times$24$\times$8.
 
On top of the backbone, two streams are used. The first stream, also known as global stream, appends a global average pooling layer to obtain a 2048-dimensional feature vector. Then, a 1$\times$1 convolution layer is used to further reduce the dimensions. The second stream, named as Batch DropBlock, \textit{randomly} removes regions on training batches. We denote this dropping module as Batch DropBlock. Then a global maximum pooling layer is appended by creating a 2048-dimensional feature vector. A maximum pooling helps to dismiss the effect of dropped regions. Finally, a fully connected layer is used to reduce the feature vector to 1024 dimensions.
 
Batch DropBlock is defined to remove a region of a pre-established size based on a ratio of input images. Since BDB Network~\cite{dai2019batch} reports the best results in regions with a third of height and the same width as the feature map, our Top DropBlock is defined specifically for the same scenario, this is, removing horizontal stripes.
 
\subsection{Top-DB-Net}

Our proposed network shares the same backbone as the baseline. Global, Top DropBlock and regularizer streams (Figure~\ref{fig:architecture}) are then appended. Global streams aim to extract general features directly from the backbone, following various previous approaches~\cite{quispe2019improved,dai2019batch,luo2019bag}. The Top DropBlock stream appends two BottleNeck layers~\cite{he2016deep} to the backbone stream and removes horizontal stripes from the most activated regions in order to push the network to maintain discriminability with less relevant data.
 
Given a training batch of $n$ images, the most activated (the most informative) stripes are defined for each image independently: the backbone outputs $n$ feature maps $F$ of size $c \times h \times w$, where $c$, $h$ and $w$ indicates channels, height and width respectively. We transform $F$ into an activation map $A$ based on the definition proposed by Zagoruyko \etal~\cite{zagoruyko2016paying}:
\begin{equation}
\label{equ:activation-maps}
A = \displaystyle\sum_{i=1}^c\abs{F_i}^p
\end{equation}
\noindent where $F_i$ represents every tensor slide of size $h \times w$. Assuming that $p > 1$ by definition~\cite{zagoruyko2016paying}, we will see that $p$ value is not relevant to our approach.

Based on $A$, we define the relevance $R$ of each stripe $r_j$ as the average of the values on row $j$:
\begin{equation}
\label{equ:stripe-relevance}
r_j = \frac{\displaystyle\sum_{k=1}^w A_{j, k}}{w}
\end{equation}
 
Finally, we can zero out rows with the largest $r_j$ values. We denote this module as Top DropBlock. For the dropping process, we create a binary mask TDM, named Top Drop Mask, of size $c \times h \times w$ for every feature map $F$ and apply the dot product between TDM and $G$, where $G$ is a tensor with the same size as $F$, which is the result of applying two BottleNeck layers~\cite{he2016deep} on $F$:
\begin{equation}
\label{equ:stripe-equivalence1}
\text{TDM}_{i, j, k} =
\begin{cases}
0, & \text{if}\ r_j \in \text{the largest values} \\
1, & \text{otherwise}
\end{cases}
\end{equation}
\noindent such that $1 \leq i \leq c$ and $1 \leq k \leq w$.
 
It is worth mentioning that, from Equations~\ref{equ:activation-maps} to~\ref{equ:stripe-relevance}, $r_j$ can be expressed as:
\begin{equation}
\label{equ:stripe-equivalence2}
r_j = \frac{\displaystyle\sum_{i=1}^c\sum_{k=1}^w \abs{F_{i, j, k}^p}}{w}
\end{equation}
 
Thus, the value of $p$ is not relevant because $\abs{x}^p \leq \abs{x}^{p+1}$ for every $p > 1$ and we use $r_j$ specifically for ranking.
 
Due to the $\abs{.}$ function in the $r_j$ definition, the most relevant stripes represent areas in $F$ with values besides zero, both positives and negatives. We can consider those to hold more discriminative information. By removing them, we push the network to learn to distinguish between samples with less available information, thus enhancing its capabilities to encode low discriminative regions. However, if the dropped regions are too large, Top DropBlock can create noise in $G$ due to false positives generated by removing regions that represent unique regions between different ID inputs.
 
To alleviate this problem, we propose a regularizer stream that will help maintain performance based on the multi-task principle~\cite{goodfellow2016deep}. This stream is only used in the training. It appends a global average pooling layer to $G$ and is then trained for ReID. Thus, it encourages $G$ to keep the information relevant to the ReID.
 
The loss function used for the three streams is the cross entropy with the label smoothing regularizer~\cite{szegedy2016rethinking} and triplet loss with hard positive-negative mining~\cite{hermans2017defense}. During the testing process, the output of global and Top DropBlock streams are concatenated.
 
\begin{table*}[!htb]
  \renewcommand{\arraystretch}{0.86}
  \setlength{\tabcolsep}{3.0mm}
  \centering
  \caption{Influence of Top-DB-Net streams and comparison with baseline.}
  \label{table:ablation}
  \begin{tabular}{lcccccccc}
  \toprule
  & \multicolumn{2}{c}{\textbf{Market1501}} & \multicolumn{2}{c}{\textbf{DukeMTMC-ReID}} & \multicolumn{2}{c}{\textbf{CUHK03 (L)}} & \multicolumn{2}{c}{\textbf{CUHK03 (D)}}\\
  \midrule
  \textbf{Method} & \textbf{mAP} & \textbf{rank-1} & \textbf{mAP} & \textbf{rank-1} & \textbf{mAP} & \textbf{rank-1} & \textbf{mAP} & \textbf{rank-1} \\
  \midrule
  no-drop Top-DB-Net & 84.7 $\pm$ 0.1 & 94.4 $\pm$ 0.3 & 72.7 $\pm$ 0.2 & 86.1 $\pm$ 0.3 & 70.7 $\pm$ 0.4 & 73.8 $\pm$ 0.6 & 68.4 $\pm$ 0.4 & 71.9 $\pm$ 0.3 \\
  no-reg Top-DB-Net  & 83.9 $\pm$ 0.1 & 93.9 $\pm$ 0.2 & 71.1 $\pm$ 0.2 & 86.1 $\pm$ 0.4 & 71.4 $\pm$ 0.4 & 74.6 $\pm$ 0.8 & 69.4 $\pm$ 0.4 & 73.5 $\pm$ 1.0 \\
  Top-DB-Net         & 85.8 $\pm$ 0.1 & 94.9 $\pm$ 0.1 & 73.5 $\pm$ 0.2 & 87.5 $\pm$ 0.3 & 75.4 $\pm$ 0.2 & 79.4 $\pm$ 0.5 & 73.2 $\pm$ 0.1 & 77.3 $\pm$ 0.5 \\
  Baseline                          & 85.2 $\pm$ 0.1 & 94.1 $\pm$ 0.1 & 73.2 $\pm$ 0.2 & 85.6 $\pm$ 0.3 & 72.2 $\pm$ 0.3 & 74.7 $\pm$ 0.6 & 70.3 $\pm$ 0.2 & 73.7 $\pm$ 0.4 \\
  \bottomrule
  \end{tabular}
\end{table*}
 
\section{Experimental Results}
\label{sec:results}
 
This section describes and discusses the main aspects related to implementation details, validation protocols and experimental results. An ablation study is carried out to analyze the effects of the Regularization and Top DropBlock streams on the Top-DB-Net. Then, we compare the results to our baseline and discuss the effects of our dropping top activation during the learning process. Finally, we compare our method to state-of-the-art approaches.

\subsection{Implementation Details}
 
All our experiments were conducted on a single Tesla v100 GPU. Due to this, we updated two items in the baseline code\footnote{We used author's~\cite{dai2019batch} original source code available at \url{https://github.com/daizuozhuo/batch-DropBlock-network}}: (i) we trained it with batch size of 64, instead of 128, and (ii) we reduced the learning rate by a factor of 0.5$\times$ because of the ``linear scaling rule''~\cite{goyal2017accurate} to minimize the effects of training with smaller batch size.
 
During the training step, input images are resized to 384$\times$128 pixels and augmented by random horizontal flip, random zooming and random input erasing~\cite{ghiasi2018DropBlock}. As mentioned previously, our Top DropBlock stream removes horizontal stripes, thus width dropping ratio is 1. Following our baseline configuration, we use a height drop ratio of 0.3. During the testing step, no drop is applied.
 
Top-DropDB-Net follows the same training setup than our baseline, based on Adam Optimizer~\cite{kingma2014adam} and a linear warm-up~\cite{goyal2017accurate} in the first 50 epochs with initial value of $1e-3$, then decayed to $1-e4$ and $1e-5$ after 200 and 300 epochs, respectively. The training routine takes 400 epochs. Due to the randomness of the drop masks used in our baseline and the methods used for data augmentation, we performed each experiment 5 times and reported the mean and standard deviation. This will allow for a fairer comparison between our method, baseline and ablation pipelines.

To combine cross entropy loss with label smoothing regularizer~\cite{szegedy2016rethinking} and triplet loss with hard positive-negative mining~\cite{hermans2017defense}, we used the neck method~\cite{luo2019bag}.
 
\begin{figure}[!htb]
  \captionsetup[subfigure]{labelformat=empty}
  \centering
  \subfloat{\includegraphics[width=1.3cm, height=2.3cm]{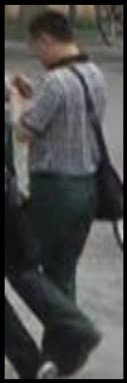}} \hspace*{0.1cm}
  \subfloat[120]{\includegraphics[width=1.3cm, height=2.3cm]{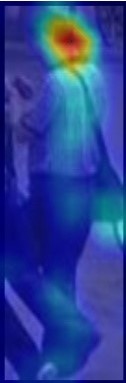}} \hspace*{0.1cm}
  \subfloat[240]{\includegraphics[width=1.3cm, height=2.3cm]{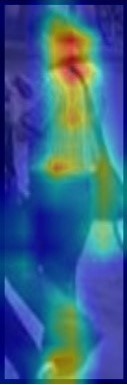}} \hspace*{0.1cm}
  \subfloat[400]{\includegraphics[width=1.3cm, height=2.3cm]{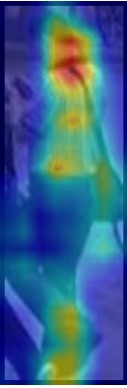}} \\
  \subfloat{\includegraphics[width=1.3cm, height=2.3cm]{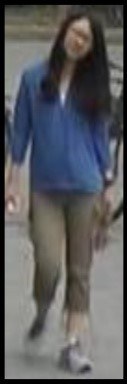}} \hspace*{0.1cm}
  \subfloat[120]{\includegraphics[width=1.3cm, height=2.3cm]{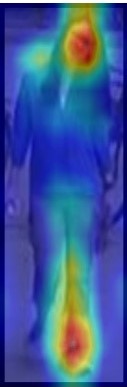}} \hspace*{0.1cm}
  \subfloat[240]{\includegraphics[width=1.3cm, height=2.3cm]{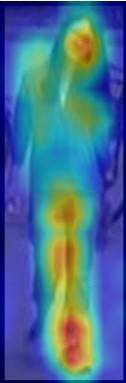}} \hspace*{0.1cm}
  \subfloat[400]{\includegraphics[width=1.3cm, height=2.3cm]{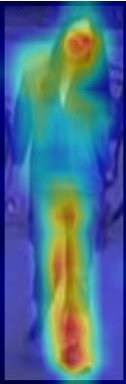}}
  \caption{Activation maps for Top-DB-Net in two images at different epochs. The number below the images indicates the epoch.}
  \label{fig:activation-examples}
\end{figure}
 
\subsection{Datasets}
 
We evaluate our framework on three widely used datasets. DukeMTMC-ReID dataset~\cite{zheng2017unlabeled,ristani2016MTMC} has hand-drawn bounding boxes with various backgrounds of outdoor scenes. Market1501 dataset~\cite{zheng2015scalable} aims to simulate a more real-world scenario and was generated through the Deformable Part Model (DPM)~\cite{felzenszwalb2009object}. CUHK03 dataset~\cite{li2014deepReID} exhibits recurrently missing body parts, occlusions and misalignment; we tested its two versions: detected (CUHK03 (D)) and labeled (CUHK03 (L)).

For training and testing, we follow the standard train/test split proposed by the dataset authors~\cite{zheng2017unlabeled,ristani2016MTMC,felzenszwalb2009object, li2014deepReID}. In the case of CUHK03, we use the new partition~\cite{zhong2017re} of 767/700, which makes this dataset more challenging. Results for each dataset are based on mean Average Precision (mAP) and Cumulative Matching Curve (CMC), more specifically, rank-1.

\subsection{Ablation Study}
 
We evaluate the effects of Top DropBlock and Regularization streams. Furthermore, we discuss the effects of Top DropBlock during the learning process and compare it to our baseline.
 
\subsubsection{\textbf{Influence of the Top DropBlock Stream}}
 
In this section, we aim to analyze the effect of our Top DropBlock stream. We train Top-DB-Net by removing the DropBlock stream and maintaining the global and regularization streams using the two Bottleneck layers. We refer to this version as \textit{no-drop Top-DB-Net}. During testing, we concatenate the output of global and regularization branches because both streams are trained with the same loss function. Results for this comparison are shown in Table~\ref{table:ablation}.
 
In all datasets, we can see that removing the Top DropBlock stream decreases performance, which is also true for the standard deviation. On the Market1501~\cite{zheng2015scalable} and DukeMTMC-ReID~\cite{zheng2017unlabeled,ristani2016MTMC} datasets, the difference is usually less than 1\% for mAP and rank-1. However, on the CUHK03~\cite{li2014deepReID,zhong2017re} dataset, we can observe significant differences: performance decreases 4.7\% in mAP and 5.6\% in rank-1 on CUHK03(L) and decreases 4.8\% in mAP and 5.4\% in rank-1 on CUHK03(D). The difference in effects between datasets may be related to the fact that CUHK03 is a more challenging benchmark. This same pattern is repeated when analyzing the effect of our Regularization stream and baseline.
 
These results are expected because the global and regularization streams follow the same optimization logic: push the backbone to encode relevant ReID information from the input images. On the other hand, when we use our Top DropBlock stream, we further encourage the backbone to recognize relevant regions and learn to describe less informative regions with richer features.
 
\subsubsection{\textbf{Influence of the Regularization Stream}}
 
In this section, our goal is to show that the regularization stream, in fact, helps to deal with the noise generated by the dropping step. For this purpose, we train a version of the Top-DB-Net without this stream, named \textit{no-reg Top-DB-Net} and compare it to the proposed Top-DB-Net. We can see in Table~\ref{table:ablation} a clear difference when using the regularization stream.
 
Using our regularization stream, we observe improvements of 1.9\% and 1\% for mAP and rank-1, respectively, on Market1501 dataset~\cite{zheng2015scalable}. DukeMTMC-ReID~\cite{zheng2017unlabeled,ristani2016MTMC} also shows improvements of 2.4\% for mAP and 1.4\% for rank-1. Similar to previous ablation analysis, the most substantial changes are for CUHK03~\cite{li2014deepReID,zhong2017re}: we can observe improvements of 4.8\% for rank-1 and 4\% for mAP on CUHK03(L), and 3.8\% for rank-1 and mAP on CUHK03(D).
 
\subsubsection{\textbf{Random DropBlock vs Top DropBlock}}
 
\begin{figure}[!htb]
  \centering
  \subfloat[Random DropBlock (Baseline)]{\includegraphics[width=5.6cm]{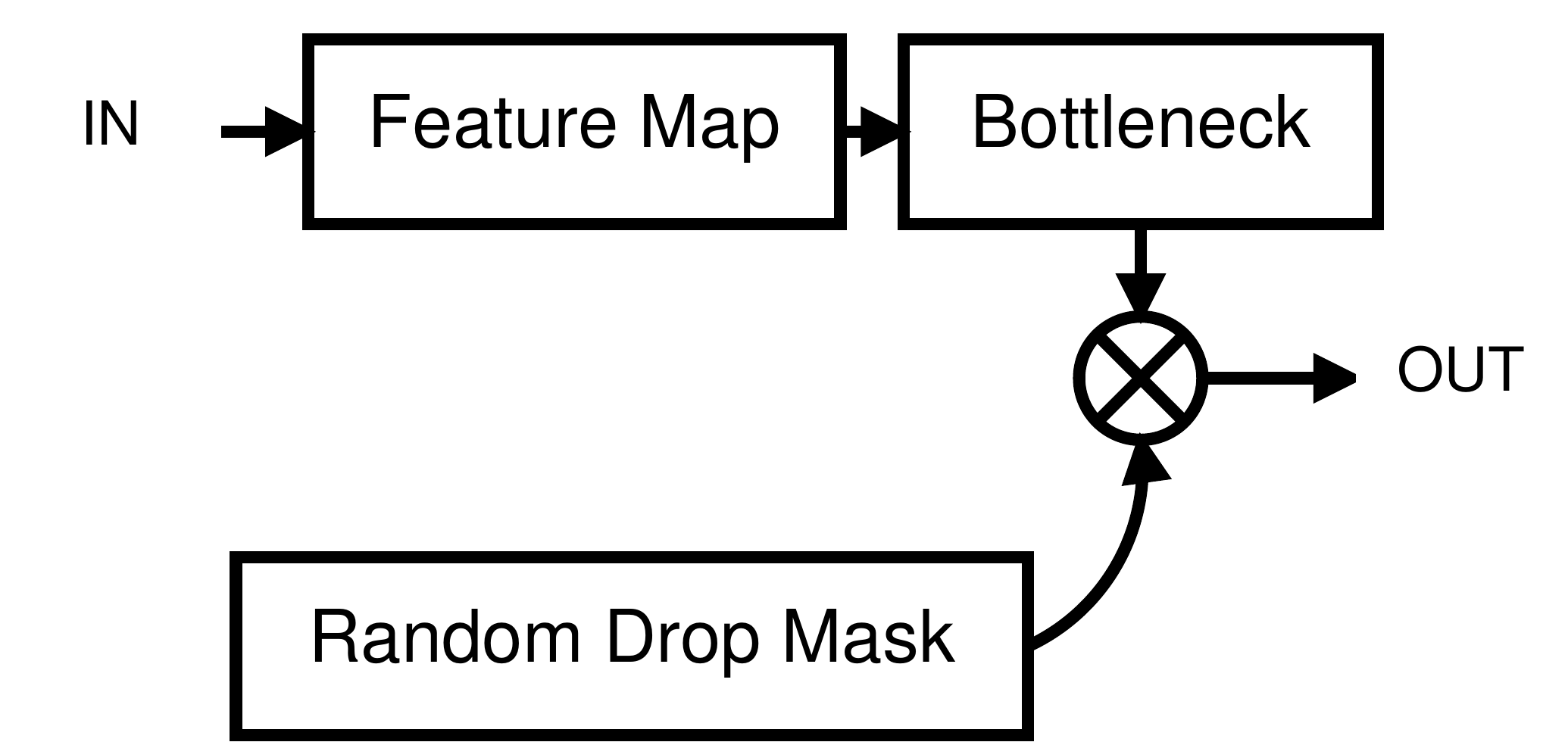}} \\
  \subfloat[Top DropBlock (Ours)]{\includegraphics[width=5.6cm]{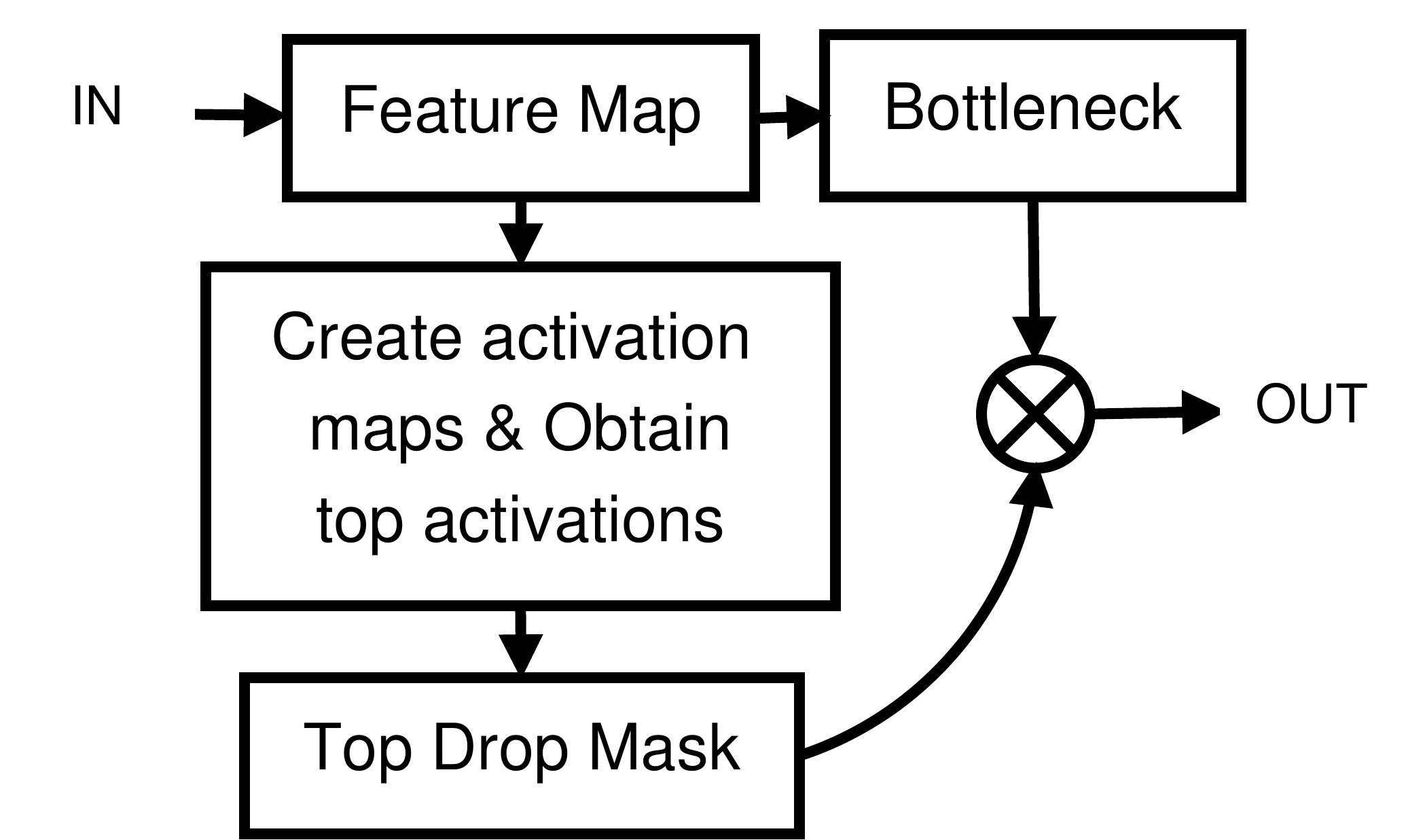}}
  \caption{Differences between the Batch DropBlock and proposed Top DropBlock.}
  \label{fig:DropBlock_vs_topDropBlock}
\end{figure}

Results in Table~\ref{table:ablation} show that our Top-DB-Net is better than our baseline in almost all metrics. The only metric with similar performance is mAP for DukeMTMC-ReID. The biggest differences are again on CUHK03~\cite{li2014deepReID,zhong2017re} dataset, with up to 4.7\% improvement for rank-1 and 3.2\% for mAP when using our Top-DB-Net. To further understand the difference in performance, we explore activation maps and their relation with the core of our method and the baseline: DropBlocks. Figure~\ref{fig:DropBlock_vs_topDropBlock} shows the differences between the two dropping methods.
 
\begin{figure*}[!htb]
  \centering
  query\hspace{0.9cm} rank-1 \hspace{0.1cm} rank-2 \hspace{0.1cm} rank-3 \hspace{0.5cm} query\hspace{0.9cm} rank-1 \hspace{0.1cm} rank-2 \hspace{0.1cm} rank-3 \hspace{0.5cm} query\hspace{0.9cm} rank-1 \hspace{0.1cm} rank-2 \hspace{0.1cm} rank-3
  \subfloat[Baseline epoch 120]{\includegraphics[width=0.3\linewidth, height=4.5cm]{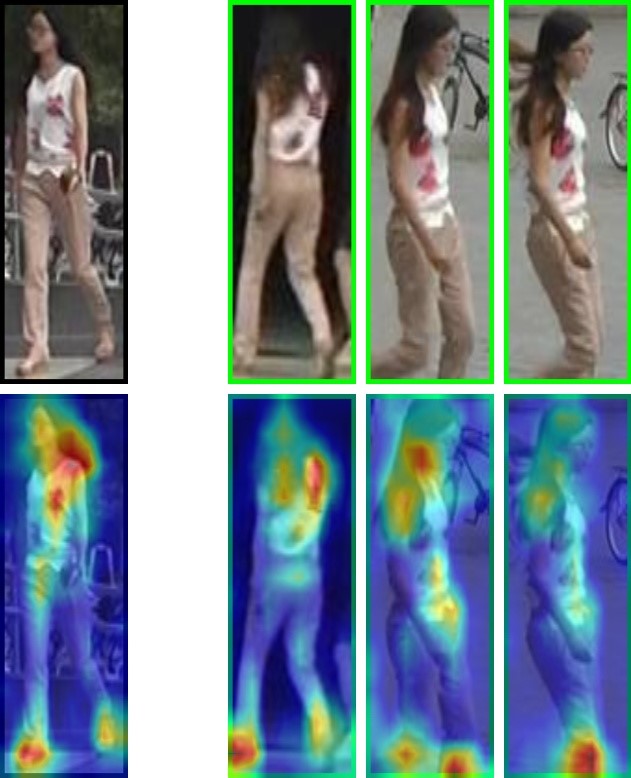}} \hspace*{0.5cm}
  \subfloat[Baseline epoch 240]{\includegraphics[width=0.3\linewidth, height=4.5cm]{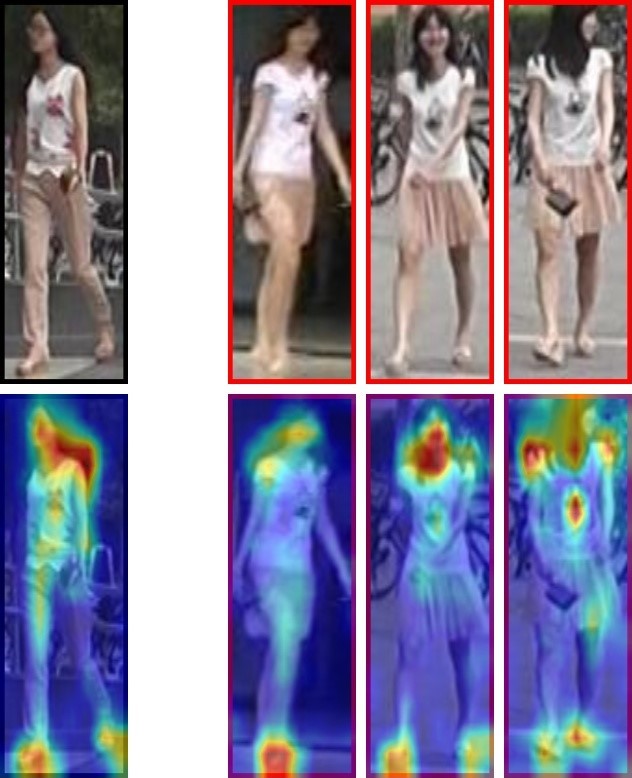}} \hspace*{0.5cm}
  \subfloat[Baseline epoch 400]{\includegraphics[width=0.3\linewidth, height=4.5cm]{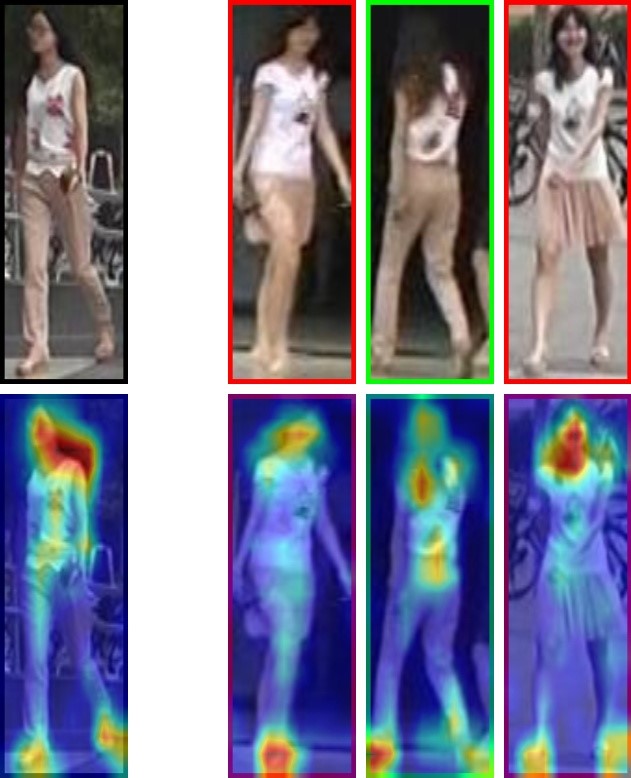}} \\[0.2cm]
  query\hspace{0.9cm} rank-1 \hspace{0.1cm} rank-2 \hspace{0.1cm} rank-3 \hspace{0.5cm} query\hspace{0.9cm} rank-1 \hspace{0.1cm} rank-2 \hspace{0.1cm} rank-3 \hspace{0.5cm} query\hspace{0.9cm} rank-1 \hspace{0.1cm} rank-2 \hspace{0.1cm} rank-3
  \subfloat[{Top-DB-Net} epoch 120]{\includegraphics[width=0.3\linewidth, height=4.5cm]{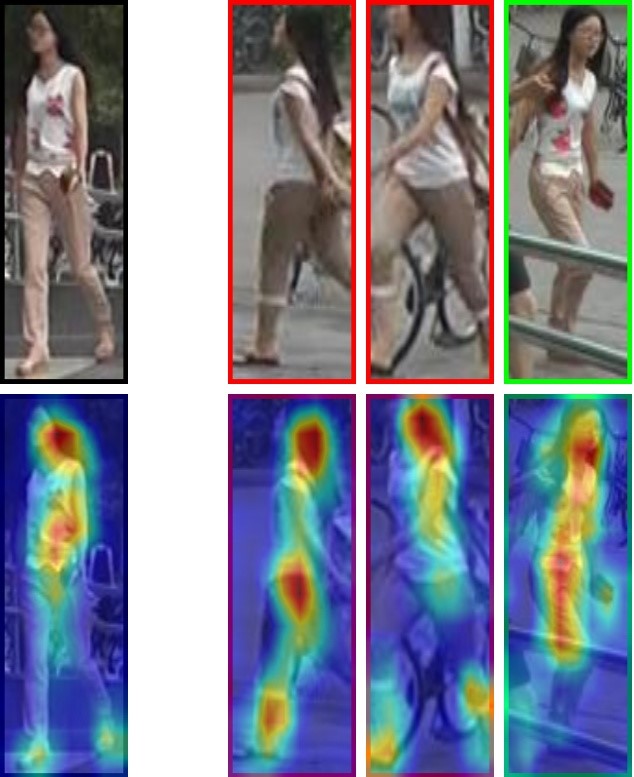}} \hspace*{0.5cm}
  \subfloat[{Top-DB-Net} epoch 240]{\includegraphics[width=0.3\linewidth, height=4.5cm]{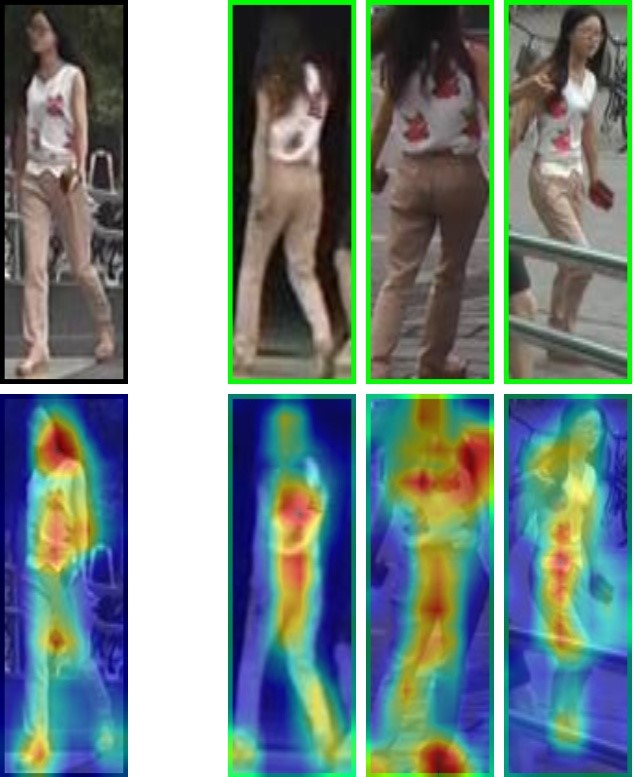}} \hspace*{0.5cm}
  \subfloat[{Top-DB-Net} epoch 400]{\includegraphics[width=0.3\linewidth, height=4.5cm]{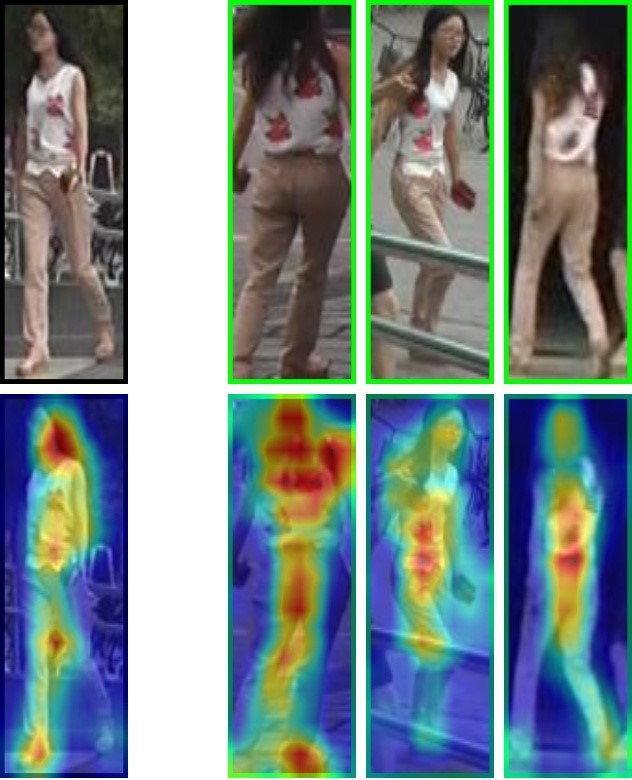}} 
  \caption{Comparison of activation and rank-3 evolution. The top and bottom sets show images for our baseline and our proposed method, respectively. We can see that using Top DropBlock, instead of Random DropBlock, makes the activations more spread out over the person, which helps to create a better feature representation. Correct results are highlighted in green, whereas incorrect results are highlighted in red.}
  \label{fig:activation-epoch}
\end{figure*}
 
In Figure~\ref{fig:activation-epoch}, we compare the evolution of rank-3 at different epochs for our Top-DB-Net and baseline. We also show the evolution of the activation maps. This example shows that our Top DropBlock improves the dispersion of activation maps in the foreground and the feature extraction from images. At 120th epoch, the activations of the query are similarly spread over the upper body and feet, both in the baseline and our method. However, we can see that, in the gallery, our method is better spread across the lower body, causing Top-DB-Net to incorrectly obtain rank-1/2, confusing a person who shares pants similar to the query. At 240th epoch, we can see that the baseline activations for the query have barely changed. Moreover, because it focuses only on the upper-body and feet, it is confused with images of a person with a similar upper body, but wearing a squirt with similar color instead of pants. On the contrary, our Top-DB-Net changed its activations for the query between 120th and 240th epochs, and also focuses on the lower body.

For this specific example, this is because our Top DropBlock removes the upper body regions and pushes the backbone to learn from the lower body since the 120th epoch, which helps to correctly match rank-1/2/3. It is also possible to notice that, because our Top DropBlock pushes the network to describe low informative regions with rich features, at 240th epoch, our network has better features to describe lower body regions, so it fixes rank-1/2 errors of 120th epoch. Finally, at 400th epoch, the baseline has still changed very little the distribution of its activations and still focuses only on the upper body and feet. It is able to obtain correct rank-2. On the other hand, our method still focuses on the entire body, retrieves the same correct baseline rank-2 and offers more similarity\footnote{Shortest Euclidean distance between features from query and gallery images.} to two images that show a strong viewpoint change and occlusions. This shows the improvement of the feature discriminability between 240th and 400th epochs.

\begin{table*}[!htb]
\renewcommand{\arraystretch}{0.86}
  \setlength{\tabcolsep}{2.5mm}
  \centering
  \caption{Comparison to the state-of-the-art approaches. RK stands for re-ranking~\cite{zhong2017re}. The sub-index indicates the ordinal position of this result (for instance, $x_3$ indicates that $x$ is the third best result).}
  \label{table:comparison-state-of-art}
  \begin{tabular}{lllllllll}
  \toprule
  & \multicolumn{2}{c}{\textbf{Market1501}} & \multicolumn{2}{c}{\textbf{DukeMTMC-ReID}}& \multicolumn{2}{c}{\textbf{CUHK03 (L)}} & \multicolumn{2}{c}{\textbf{CUHK03 (D)}} \\
  \midrule
  \textbf{Method} & \textbf{mAP} & \textbf{rank-1} & \textbf{mAP} & \textbf{rank-1} & \textbf{mAP} & \textbf{rank-1} & \textbf{mAP} & \textbf{rank-1} \\
  \midrule
  BoT~\cite{luo2019bag}                 & 85.9$_5$ & 94.5     & 76.4     & 86.4     &  --      &  --      &  --      &  --      \\
  PyrNet~\cite{martinel2019aggregating} & 86.7     & 95.2$_3$ & 74.0     & 87.1     & 68.3     & 71.6     & 63.8     & 68.0     \\
  Auto-ReID~\cite{quan2019auto}         & 85.1     & 94.5     &  --      &  --      & 73.0     & 77.9     & 69.3     & 73.3     \\
  MGN~\cite{wang2018learning}           & 86.9$_4$ & 95.7$_1$ & 78.4$_3$ & 88.7$_4$ & 67.4     & 68.0     & 66.0     & 66.8     \\
  DenSem~\cite{Zhang_2019_CVPR}         & 87.6$_3$ & 95.7$_1$ & 74.3     & 86.2     & 75.2     & 78.9     & 73.1     & 78.2$_3$ \\
  IANet~\cite{hou2019interaction}       & 83.1     & 94.4     & 73.4     & 87.1     &  --      &  --      &  --      & --       \\
  CAMA~\cite{yang2019towards}           & 84.5     & 94.7     & 72.9     & 85.8     &  --      &  --      &  --      & --       \\
  MHN~\cite{chen2019mixed}              & 85.0     & 95.1$_4$ & 77.2     & 89.1$_2$ & 72.4     & 77.2     & 65.4     & 71.7     \\
  ABDnet~\cite{chen2019abd}             & 88.2$_2$ & 95.6$_2$ & 78.5$_2$ & 89.0$_3$ &  --      &  --      &  --      & --       \\
  SONA~\cite{xia2019second}             & 88.6$_1$ & 95.6$_2$ & 78.0     & 89.2$_1$ & 79.2$_1$ & 81.8$_1$ & 76.3$_1$ & 79.1$_1$ \\
  OSNet~\cite{zhou2019osnet}            & 84.9     & 94.8     & 73.5     & 88.6$_5$ &  --      &  --      & 67.8     & 72.3     \\
  Pyramid~\cite{zheng2019pyramidal}     & 88.2$_2$ & 95.7$_1$ & 79.0$_1$ & 89.0$_3$ & 76.9$_2$ & 78.9     & 74.8$_2$ & 78.9$_2$ \\
  Top-DB-Net (Ours)                    & 85.8$_6$ & 94.9$_5$ & 73.5     & 87.5$_6$ & 75.4$_3$ & 79.4$_2$ & 73.2$_3$ & 77.3$_4$ \\
 
  \midrule
  SSP-ReID+RK~\cite{quispe2019improved}    & 90.8     & 93.7     & 83.7     & 86.4     & 77.5     & 74.6     & 75.0     & 72.4     \\
  BoT+RK~\cite{luo2019bag}                 & 94.2$_1$ & 95.4     & 89.1$_1$ & 90.3$_2$ &  --      &  --      &  --      &  --      \\
  PyrNet+RK~\cite{martinel2019aggregating} & 94.0     & 96.1$_1$ & 87.7     & 90.3$_2$ & 78.7$_2$ & 77.1$_2$ & 82.7$_2$ & 80.8$_2$ \\
  Auto-ReID+RK~\cite{quan2019auto}         & 94.2$_1$ & 95.4     &  --      &  --      &  --      &  --      &  --      &  --      \\
  Top-DB-Net+RK (Ours)                     & 94.1$_2$ & 95.5$_2$ & 88.6$_2$ & 90.9$_1$ & 88.5$_1$ & 86.7$_1$ & 86.9$_1$ & 85.7$_1$ \\
  
  \bottomrule
  \end{tabular}
\end{table*}

Activation plots are useful for the interpretability of networks. In our case, our plots are generated following Equation~\ref{equ:activation-maps}. This equation is also used to define our Top DropBlock and Top Drop masks. This shows that the same tool used for interpretability can also be applied during the learning process to enhance discriminability (Table~\ref{table:ablation}). In addition to quantitative improvements, we observe a clear improvement in the quality of regions where backbone is concentrated. As shown in Figures~\ref{fig:activation-examples} and~\ref{fig:activation-epoch}, there is a consistent and significant activation improvement between 120th and 240th epochs, when they start to focus on broader body parts. From 240th to 400th epochs, we can see that the activations become more stable and well spread out in the foreground, but with an enhanced discriminability. We use the activation definition of Zagoruyko \etal~\cite{zagoruyko2016paying} because it adjusts to our network pipeline and drop objective. However, there is a previous work for ReID~\cite{yang2019towards} that uses CAM~\cite{zhou2016learning}, an activation definition that introduces weights for each channel to enhance the scope of network activation. Previous literature and our findings suggest that methods used for interpretability may be useful to improve ReID and network activation in general.
 
\subsection{State-of-the-Art Comparison}
 
Our method focuses on ReID using information extracted only from input images. Thus, in our comparison to the state-of-the-art, we consider methods in a similar way, for instance, Zhu\etal~\cite{zhu2020aware} used the camera ID and Wang\etal~\cite{wang2019spatial} used the image time-stamp during training. This extra information may bias the models to learn the mapping between the camera and views or the time needed for a person to move from different viewpoints, instead of extracting reliable information from images, so that they are not included in our comparison. Table~\ref{table:comparison-state-of-art} shows a comparison between our method and state-of-the-art approaches. We compare the results separately when using re-ranking~\cite{zhong2017re}.

Our results are among the top-6 results for both mAP and rank-1 on Market1501. We have a similar performance for rank-1 on DukeMTMC-ReID, however, for mAP, we achieved results comparable to state-of-the-art methods, such as OSNet~\cite{zhou2019osnet}, CAMA~\cite{yang2019towards} and IANet~\cite{hou2019interaction}. We obtained the second best rank-1 on CUHK03(L), third best mAP on both versions of CUHK03 and fourth best rank-1 on CUHK03(D). When using re-ranking, our method achieved state-of-the-art results on CUHK03(L) and CUHK03(R) in both mAP and rank-1, as well as best results for rank-1 on DukeMTMC-ReID, second best mAP on DukeMTMC-ReID and second best on Market1501 in both mAP and rank-1.

\section{Conclusions and Future Work}
\label{sec:conclusions}

In this paper, we introduced Top-DB-Net, a network for the person re-identification problem based on Top DropBlock. Top-DB-Net encourages the network to improve its performance by learning to generate rich encoding based on low informative regions. It consists of three streams: a global stream that follows standard feature encoding for the backbone, the Top DropBlock that pushes the network to maintain its performance while using less discriminative regions by dropping most activated parts of the feature map, and a regularization stream that helps to deal with the noise created by the dropping process.

Extensive experiments conducted on three widely datasets demonstrated the power of our method to achieve competitive results and its capability to  generate better activation maps than competing methods. Moreover, our results suggest that methods proposed for interpretability of activation maps can help during training in ReID.

As directions for future work, we expect to extend the definition of Top DropBlock to various dropping ratios, instead of only horizontal stripes. Furthermore, we intend to encode higher level features (for instance, gender) and analyze their impact on the ReID task.
 
\section*{Acknowledgments}
\label{acknowledgment}
 
Part of this work was done while the first author was affiliated with Microsoft Corp. We are thankful to Microsoft Research, S\~ao Paulo Research Foundation (FAPESP grant \#2017/12646-3), National Council for Scientific and Technological Development (CNPq grant \#309330/2018-1) and Coordination for the Improvement of Higher Education Personnel (CAPES) for their financial support.
 
\balance
\bibliographystyle{IEEEtran}
\bibliography{paper}
 
\end{document}